\documentclass[10pt,sigconf,nonacm]{acmart}

\AtBeginDocument{%
  \providecommand\BibTeX{{%
    \normalfont B\kern-0.5em{\scshape i\kern-0.25em b}\kern-0.8em\TeX}}}

\usepackage{xspace}
\newcommand{\MIRAGEA}{$\mathtt{MIRAGE}19$\xspace}

\newcommand{\noaug}{$\text{No Aug}$\xspace}
\newcommand{\noaugwos}{$\text{No Aug w/o Sampler}$\xspace}

\newcommand{\FlipTime}{$\text{Flip}$\xspace}
\newcommand{\ShiftTime}{$\text{Translation}$\xspace}
\newcommand{\Permutation}{$\text{Permutation}$\xspace}
\newcommand{\InterpolateTime}{$\text{Interpolation}$\xspace}
\newcommand{\TimeWrap}{$\text{Wrap}$\xspace}
\newcommand{\PacketLossIMC}{$\text{Packet Loss}$\xspace}

\newcommand{\ChannelNoiseGaussian}{$\text{Gaussian Noise}$\xspace}
\newcommand{\ChannelNoiseSpike}{$\text{Spike Noise}$\xspace}
\newcommand{\ManitudeNormalWrapUp}{$\text{Gaussian WrapUp}$\xspace}
\newcommand{\ManitudeSinWrap}{$\text{Sine WrapUp}$\xspace}
\newcommand{\MaskWindow}{$\text{Window Mask}$\xspace}
\newcommand{\MaskBernoulli}{$\text{Bernoulli Mask}$\xspace}
\newcommand{\ChangeRTTIMC}{$\text{Constant WrapUp}$\xspace}

\newcommand{\CutMix}{$\text{CutMix}$\xspace}

\usepackage{setspace}
\usepackage{booktabs}
\usepackage{xcolor}
\usepackage{cellspace}
\usepackage{wasysym}
\newcommand{\circleempty}{\textcolor{gray}{\Circle}}
\newcommand{\circlefull}{\textcolor{gray}{\CIRCLE}}
\newcommand{\circlesemi}{\textcolor{gray}{\LEFTcircle}}

\usepackage{array, makecell, multirow,rotating}

\usepackage{soul}
\usepackage{natbib}
\usepackage{lipsum}

\begin{document}

\title[]{
    \fontsize{17}{17}\selectfont
    Toward Generative Data Augmentation for Traffic Classification
}

\newcommand{\afnote}[1]{\textcolor{red}{AF:#1}}

\author{Chao Wang}
\affiliation{
  \institution{\mbox{Huawei Technologies SASU, France}}
  \country{}
}
\email{wang.chao3@huawei.com}

\author{Alessandro Finamore}
\affiliation{
 \institution{\mbox{Huawei Technologies SASU, France}}
  \country{}
}
\email{alessandro.finamore@huawei.com}

\author{Pietro Michiardi}
\affiliation{
  \institution{Eurecom, France}
  \country{}
}
\email{pietro.michiardi@eurecom.fr}

\author{Massimo Gallo}
\affiliation{
   \institution{\mbox{Huawei Technologies SASU, France}}
  \country{}
}
\email{massimo.gallo@huawei.com}

\author{Dario Rossi}
\affiliation{
   \institution{\mbox{Huawei Technologies SASU, France}}
  \country{}
}
\email{dario.rossi@huawei.com}

\renewcommand{\shortauthors}{}

\begin{abstract}
Data Augmentation (DA)---augmenting training data with synthetic 
samples---is wildly adopted in Computer Vision (CV) 
to improve models performance. 
Conversely, DA has not been yet popularized in networking use cases, including Traffic Classification (TC).
In this work, we present a preliminary study 
of $14$ hand-crafted DAs applied on the \MIRAGEA dataset.
Our results ($i$) show that DA can reap benefits
previously unexplored in TC and ($ii$) 
foster a research agenda on the use of \emph{generative models} to automate DA design.
\end{abstract}

\begingroup
\mathchardef\UrlBreakPenalty=10000
\maketitle
\endgroup

\section{Introduction}

Network monitoring is at the core of networks operations with
Traffic Classification (TC) being key for traffic management.
Traditional Deep Packet Inspection (DPI) techniques, i.e., classifying traffic
with rules related to packets content, is nowadays challenged by the growth in adoption of TLS/DNSSEC/HTTPS encryption.
Despite the quest for alternative solutions to DPI already sparked three
decades ago with the first Machine Learning (ML) models based on packet and flow features, 
a renewed thrust
is fueled today by the rise of Deep Learning (DL),
with abundant TC literature reusing/adapting Computer Vision (CV) training algorithms and 
model architectures~\cite{dl4tc2}.

In this work, 
we argue that \emph{opportunities laying in the data itself are still underexplored}, 
based on two observations.
First, CV and Natural Language Processing (NLP)
adopt  ``cheap'' \emph{Data Augmentation (DA)} 
strategies (e.g., image rotation or
synonym replacement)
for 
improving models performance.
Yet, almost no TC literature investigates DA.
Second, 
\emph{network traffic datasets are imbalanced in nature} 
due to 
app/service popularity skew,
which calls for strategies to augment the minority classes.
Again, the interplay between imbalance and model performance
is typically ignored in TC literature.

In this paper, we propose a two-fold research agenda: 
($a$) first, we study 
hand-crafted DA
to assess its benefits and relationship with class imbalance (Sec.~\ref{sec: Hand-crafted Augmentation});
then, ($b$) we charter
a roadmap to pursue better augmentation strategies via \emph{generative models}, i.e., 
learning DA in a data-driven fashion
rather than adopting manual design (Sec.~\ref{sec: Generation-based Augmentation}).

\section{Hand-crafted Augmentations}\label{sec: Hand-crafted Augmentation}
We define hand-crafted DA as the family of transformations that can be described by simple mathematical formulations or algorithms, 
e.g., additive noise, random masking, and interpolation between samples (just to name a few). 
Such transformations are meant to be \textit{directly} used in the \textbf{input space}, thus 
their design requires domain knowledge 
to control samples variety---too little 
produces simple duplicates; too much breaks class semantics and introduces undesired data shifts. 
At the same time, such DA aims at \emph{indirectly} modifying 
DL classifiers decision boundaries
in the \textbf{latent space}. Indeed, to be beneficial, 
DA should introduce additional training points 
that foster better clustering of the classes in the latent space. 
However, without explicitly knowledge of how input samples are projected in the latent space (as models are ``black boxes''), 
domain knowledge 
hardly suffices for effectively designing these 
transformations---the use of DA is a trial and error process.
Moreover, even if many studies investigated DA for CV and
time series (e.g., in the medical field), 
reusing such 
methods is not trivial for TC as data suffers 
from two extra undesirable restrictions:
\emph{input samples are short}---traditionally, they are time series of the first N packets of a flow (e.g., the first $10$ packet sizes and inter arrival times)\footnote{Focusing on the first packets is required to enable early classification and effectively enforce traffic management policies.}
---and are
\emph{semantically weak}---interpreting packet time series is 
less obvious than interpreting electrocardiograms. 

\subsection{Our preliminary study}
Inspired by time-series DA literature~\cite{TS_aug_survey},
we formulated a preliminary empirical study based on
the \MIRAGEA\footnote{\url{https://traffic.comics.unina.it/mirage/mirage-2019.html}} dataset which gathers traffic from 20 Android 
apps. We benchmarked  $14$ augmentations 
applied on the first $20$ packets size, direction and Inter Arrival Time (IAT)
for each flow.
Each augmentation is applied 
once
to each batch of size $B$ hence doubling 
its size to $2B$.\footnote{
At the end of a training epoch a model 
has observed the original version of the dataset and a randomized version of 
all the training samples. Different epochs generate different randomized
versions of the data.}
We compared against models trained without DA 
with batches of size $2B$.
Given the imbalance, we use a class-weighted sampler which creates batches
by selecting 
minority class samples more frequently.

\begin{figure}[t]
\centering
\includegraphics[width=1\columnwidth]{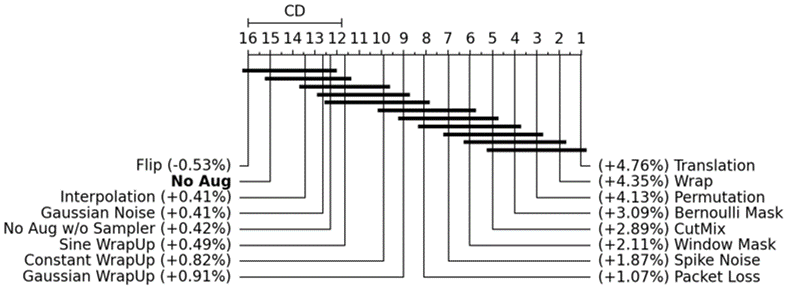}
\setstretch{0.5}
\fontsize{5}{10}\selectfont
\raggedright
\begin{tabular}{
    @{}
    l@{$\,\,$}
    l@{$\,\,$}
    l@{$\,\,$}
    l@{$\,\,$}
    l@{$\,\,$}
    p{0.7\columnwidth}
    @{}
}
\\
\toprule
 & 
\bf Aug. Name & 
\multicolumn{3}{@{}l@{$\,\,$}}{\bf Alter Feat.} & 
\bf Description
\\
& &
\it size &
\it dir &
\it iat &
\\
\midrule
& No Aug                  & \circleempty & \circleempty & \circleempty & Reference baseline
\\
& No Aug  w/o Sampler     & \circleempty & \circleempty& \circleempty & --
\\
\midrule
\parbox[t]{1.3mm}{\multirow{5}{*}{\rotatebox[origin=c]{90}{Amplitude}} } 
& Gaussian Noise          & \circlesemi & \circleempty & \circlesemi & Gaussian noise  added to all values
\\
&  Spike Noise            &\circlesemi & \circleempty & \circlesemi  &  Absolute of Gaussian noise added to at most $3$ non zero values
\\
&  Gaussian WrapUp        & \circlesemi & \circleempty & \circlesemi  &  Gaussian signal multiplied to all values
\\
&  Sine WrapUp            & \circlesemi & \circleempty & \circlesemi  &  Sinusoidal signal multiplied to all values
\\
&  Constant WrapUp        &\circleempty &\circleempty &  \circlefull &  Single uniformly sampled value multiplied to all IAT values only
\\
\midrule
\parbox[t]{1.3mm}{\multirow{2}{*}{\rotatebox[origin=c]{90}{Mask}} }
&  Bernoulli Mask        &  \circlefull & \circlefull & \circlefull & Zeroing values with predefined probability
\\
& Window Mask            & \circlefull & \circlefull & \circlefull & Zeroing a segment of size $2$ at random starting point 
\\
\midrule
\parbox[t]{1.3mm}{\multirow{7}{*}{\rotatebox[origin=c]{90}{Order}} }
&  Interpolation         & \circlefull & \circlefull & \circlefull  & Add avg. values in-between each pair, and select 20 consec. values from random start\\
&  Flip                  & \circlefull & \circlefull & \circlefull &  Swap values left to right
\\
&  Packet Loss           & \circlefull & \circlefull & \circlefull  &  Discard packets arrived in an interval $\triangle t$ and add zero padding (if needed)\\
&  Translation           & \circlefull & \circlefull & \circlefull  &  
Select a random starting point and shift values left or right
\\
&  Wrap                  &\circlefull & \circlefull & \circlefull  &  Interpolate or discard packets according to a randomized indicator
\\
&  Permutation           & \circlefull & \circlefull & \circlefull   &  Split the time series into segments of random sizes and shuffle them
\\
&  CutMix                &  \circlefull & \circlefull & \circlefull  & Create new samples by patching segments from 2 different samples\\
\bottomrule
\end{tabular}
Segment: a consecutive group of timestamps. 
A feature is altered with probability $p=1$ \circlefull, 
$p=0.5$ \circlesemi, 
or
$p=0$ \circleempty 
\vspace{-5pt}
\caption{Critical distance chart on \MIRAGEA 
(No Aug refers to weighted F1 score = 75.21\%. Difference of augmentation's metric w.r.t. baseline in brackets).
\label{fig:1_fixed_CD_MIRAGE19}
}
\end{figure}

Figure~\ref{fig:1_fixed_CD_MIRAGE19} reports the Critical Difference (CD) chart of the weighted-F1 score, across 30 seeds, for each augmentation. To create such summary, first the performance of a given seed are ranked (the smaller, the better); then the ranks across seeds are analyzed via a Friedman test with Nemenyi post-hoc test which assesses their overlap and connects
with an horizontal line augmentations that are not statistically different. 
First of all, notice how using the weighted samplers alone slightly hurts performance (\noaug 
$-0.4\%$ w.r.t. \noaugwos). This is because the accuracy for majority classes 
reduces when sampling more frequently minority classes---showing more often
the same minority samples does not help the learning.
Conversely, 
combining this sampling with DA 
yields sizable improvements (up to +4.76\% w.r.t. \noaug).
In fact, despite the higher attention toward smaller classes,
we also observe relative improvement in performance
for larger classes (results not reported for lack of space).
This hints that the samples variety added by some DAs indeed helps models to learn better data representations.

\section{Generative data augmentation}\label{sec: Generation-based Augmentation}

Studies like the one in Fig.~\ref{fig:1_fixed_CD_MIRAGE19} are
empirical and costly.\footnote{The search space expands when ``stacking''
DAs as done in CV.}
More importantly, 
implications on performance of DA methods are hard to predict.
However, we argue that
these campaigns are instrumental to charter 
the road toward \emph{automatically learning augmentations}.
In particular, we identify three stages.

\noindent \textbf{Latent space geometry.}
First of all, 
it is reasonable to assume
that the augmentations performance gaps is rooted in 
the 
geometry of the latent space.
In fact, good DAs 
encourage the learning of more general and robust features, resulting in 
better classes separation.
This raises questions
such as \emph{``Where would be more effective to project synthetic points? 
What level of samples variety is most effective for training?''}
which we will address by using clustering metrics and
latent space geometry analysis---we aim to 
uncover how augmentations can help to ``regularize'' the latent space.

\noindent \textbf{Generative models.}
Second, better DA should be viable 
via generative models such as
Generative Adversarial Networks (GAN)
and Diffusion Models (DM).
These techniques approximate the input data distribution, generate diverse samples, and can be guided by conditional mechanisms to steer their projection in the latent space of a classifier.
The high realism and diversity of the obtained synthetic samples
have motivated their use
for augmenting training datasets 
for classification tasks~\cite{dafussion}. 

However, generative models are usually trained separately 
from the final downstream task and with
datasets having a large variety of samples.
Unfortunately, state of the art datasets in TC 
offer
only a modest variety compared to CV datasets.\footnote{
CESNET-TLS22 has 191 services across 141 million flows;
LAION5B has 5 billion image-text pairs.
}
Linking back to the previous stage,
we envision a first exploration based on 
\emph{conditioning} the generative models on the latent space properties learned via hand-crafted DA.
Then, we will target the more challenging scenario of
training \emph{unconditionally} using datasets enlarged with hand-crafted DA and verify if effective regularizations are automatically learned.

\noindent{\bf Training pipeline.}
Last, we expect generative models based on pre-training
to be sub-optimal in TC due to lower variety in the data.
To link generative models to classification needs
we will consider also an end-to-end training pipeline where both classifier
and generative model are learned 
jointly.
This calls for \emph{self-supervision} mechanisms
which have already been reported as useful is recent TC studies~\cite{SSLTC}.

\section{Conclusion}

In this paper we presented a preliminary study supporting the use of DA
and outlined a research agenda for adopting generative models in TC.
We also raised awareness toward a more
careful investigation of dataset imbalance.
We believe that tackling the highlighted challenges
will bring meaningful insights to TC.

\bibliographystyle{unsrtnat}
\bibliography{camera-ready/bibliography}

\begin{thebibliography}{4}
\providecommand{\natexlab}[1]{#1}
\providecommand{\url}[1]{\texttt{#1}}
\expandafter\ifx\csname urlstyle\endcsname\relax
  \providecommand{\doi}[1]{doi: #1}\else
  \providecommand{\doi}{doi: \begingroup \urlstyle{rm}\Url}\fi

\bibitem[Salman et~al.(2020)Salman, Elhajj, Kayssi, and Chehab]{dl4tc2}
Ola Salman, Imad~H. Elhajj, Ayman Kayssi, and Ali Chehab.
\newblock A review on machine learning–based approaches for internet traffic
  classification.
\newblock In \emph{Annals of Telecommunications}, 2020.

\bibitem[Wen et~al.(2021)Wen, Sun, Yang, Song, Gao, Wang, and
  Xu]{TS_aug_survey}
Qingsong Wen, Liang Sun, Fan Yang, Xiaomin Song, Jingkun Gao, Xue Wang, and
  Huan Xu.
\newblock Time series data augmentation for deep learning: A survey.
\newblock In \emph{IJCAI}, 2021.

\bibitem[Trabucco et~al.(2023)Trabucco, Doherty, Gurinas, and
  Salakhutdinov]{dafussion}
Brandon Trabucco, Kyle Doherty, Max Gurinas, and Ruslan Salakhutdinov.
\newblock Effective data augmentation with diffusion models, 2023.

\bibitem[Towhid and Shahriar(2022)]{SSLTC}
Md.~Shamim Towhid and Nashid Shahriar.
\newblock Encrypted network traffic classification using self-supervised
  learning.
\newblock In \emph{NetSoft}, 2022.

\end{thebibliography}

\end{document}